\documentclass[letterpaper, 10 pt, conference]{ieeeconf}  

\IEEEoverridecommandlockouts                              
\usepackage{graphics} 
\usepackage{amsmath} 
\usepackage{amssymb}  
\usepackage{graphicx}
\usepackage{subcaption}
\usepackage{multirow}
\usepackage[table,xcdraw]{xcolor}
\usepackage{tabularx}

\usepackage[breaklinks,colorlinks]{hyperref}

\title{\LARGE \bf
Light-Weight Vision Transformer\\with Parallel Local and Global Self-Attention
}

\author{Nikolas Ebert$^{1,2}$, Laurenz Reichardt$^{1}$, Didier Stricker$^{2}$ and Oliver Wasenm\"uller$^{1}$
\thanks{$^{1}$Mannheim University of Applied Sciences, Germany}
\thanks{{\tt\small n.ebert@hs-mannheim.de}}
\thanks{{\tt\small l.reichardt@hs-mannheim.de}}
\thanks{{\tt\small o.wasenmueller@hs-mannheim.de}}
\thanks{$^{2}$RPTU Kaiserslautern-Landau, Germany}
\thanks{{\tt\small didier.stricker@dfki.de}}
}

\begin{document}

\maketitle
\thispagestyle{empty}
\pagestyle{empty}

\begin{abstract}

While transformer architectures have dominated computer vision in recent years, these models cannot easily be deployed on hardware with limited resources for autonomous driving tasks that require real-time-performance. Their computational complexity and memory requirements limits their use, especially for applications with high-resolution inputs. In our work, we redesign the powerful state-of-the-art Vision Transformer PLG-ViT to a much more compact and efficient architecture that is suitable for such tasks. We identify computationally expensive blocks in the original PLG-ViT architecture and propose several redesigns aimed at reducing the number of parameters and floating-point operations. As a result of our redesign, we are able to reduce PLG-ViT in size by a factor of 5, with a moderate drop in performance. We propose two variants, optimized for the best trade-off between parameter count to runtime as well as parameter count to accuracy. With only 5 million parameters, we achieve 79.5$\%$ top-1 accuracy on the ImageNet-1K classification benchmark. Our networks demonstrate great performance on general vision benchmarks like COCO instance segmentation. In addition, we conduct a series of experiments, demonstrating the potential of our approach in solving various tasks specifically tailored to the challenges of autonomous driving and transportation. 

\end{abstract}

\section{Introduction}

The emergence of vision transformers (ViTs) \cite{dosovitskiy2020image, liu2021swin, wang2021pyramid} as a viable alternative to convolutional neural networks (CNNs) \cite{he2016deep} stems from the success of the multihead self-attention mechanisms \cite{vaswani2017attention}, which, in contrast to standard convolution, provide a global receptive field. However, while ViTs have demonstrated their potential in various visual recognition tasks, the computational complexity of self-attention presents a challenge in adapting these methods to resource constrained applications.
\begin{figure}[t]
\centering
\subfloat{\includegraphics[width = 0.99\linewidth]{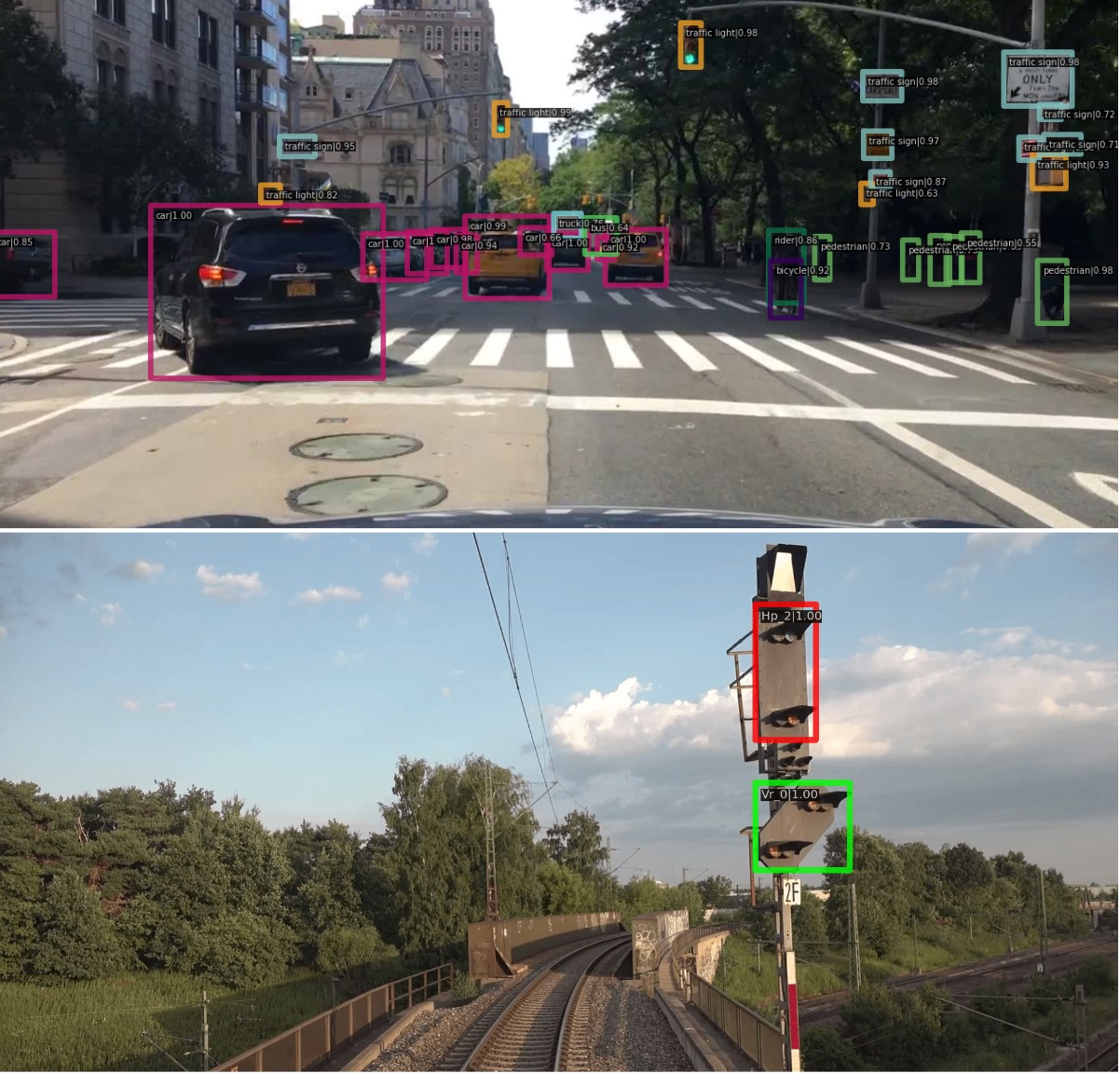}}\\
\caption{Example detections of our Light-Weight PLG-ViT backbone within a Faster-RCNN \cite{Ren_2017} on the datasets BDD100k \cite{yu2020bdd100k} and GERALD \cite{leibner2023gerald}.}
\label{fig:cover}
\vspace{-4mm}
\end{figure}
For image processing on hardware with limited resources, especially in the field of autonomous driving, transformer-models with large amount of parameters and high computational complexity are not suitable for tasks which require realtime or near-realtime performance.
Typically, lightweight CNNs are used for visual recognition tasks on resource-constrained platforms.

Recently, there have been first approaches for a lightweight transformer design such as MobileViT \cite{mehta2022mobilevit} and EdgeViT \cite{pan2022edgevits}. These networks leverage hybrid architectures combining CNNs with transformers in order to reduce complexity as well as memory and show great potential for the use of transformers for mobile applications. 
Furthermore, continued improvement of MobileViT \cite{mehta2022separable, wadekar2022mobilevitv3} shows that this potential has not yet been fully exploited.

\begin{figure*}[t]
\centering
\includegraphics[width=0.7\linewidth]{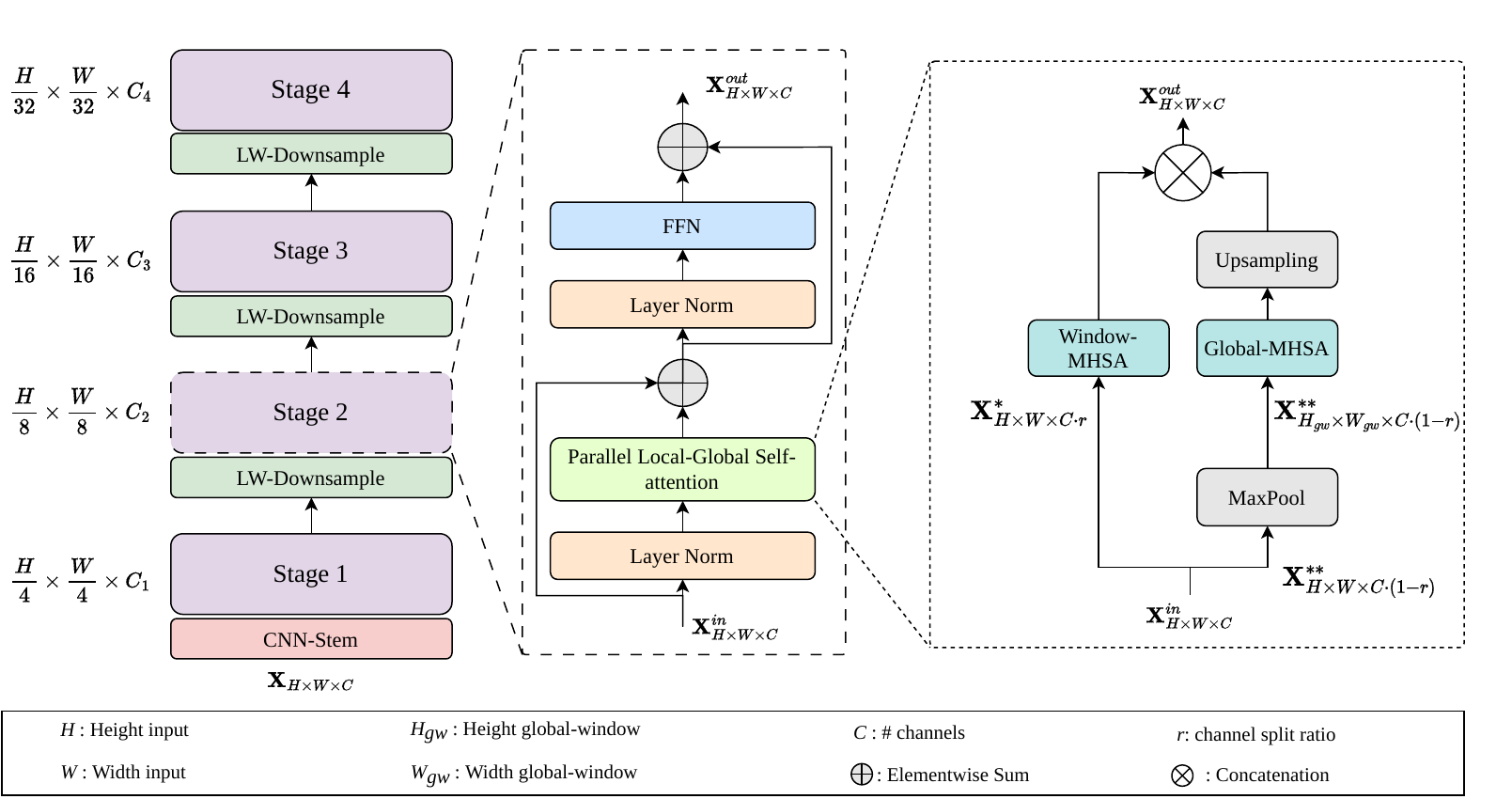}
\caption{\textit{Left:} Architecture of our LW PLG-ViT. \textit{Middle:} LW PLG-ViT block consists of the parallel local-global self-attention (SA) and our CCF-FFN+. \textit{Right:} Parallel local and global SA.}
\label{fig:architektur}
\end{figure*}

In this paper, we present our Light-Weight (LW) PLG-ViT, a computationally efficient but powerful feature extractor for use on resource-constrained hardware. We build on the general-purpose PLG-ViT \cite{ebert2023plg}, as its parallel local-global attention makes it an ideal candidate for hardware-limited platforms. Specifically, the unique concept of utilizing a resolution-adaptive window for global attention maintains a favorable trade-off between low computational complexity and high accuracy. While this specific design choice is beneficial, the rest of PLG-ViT remains resource intensive.

In a comprehensive study, we identify and replace computationally inefficiencies with our novel innovations to increase the networks computational efficiency while decreasing parameter count and memory requirements. Specifically, these innovations are a performant downsampling operation between transformer stages and a new and more powerful feed-forward network within the transformer layer.
Finally, we demonstrate the superior performance of LW PLG-ViT through a comprehensive evaluation on general vision benchmarks \cite{deng2009imagenet, lin2014microsoft} as well as four benchmarks in the area of intelligent transportation \cite{leibner2023gerald, yu2020bdd100k, caesar2020nuscenes, DiasDaCruz2020SVIRO} (Figure \ref{fig:cover}).

\section{Related Works}
\subsection{Light-Weight CNNs}
The great success of CNNs \cite{he2016deep} has naturally led to the development of various efficient architectures and factorizations of the basic convolution layer for use on restricted devices. 
Recent lightweight networks are characterized primarily by real-time processing and low energy consumption, which is crucial for resource constrained applications.
Architectures like MobileNet \cite{howard2017mobilenets, sandler2018mobilenetv2} and ShuffleNet \cite{zhang2018shufflenet} have become widely adopted in autonomous driving \cite{siam2018comparative} due to their ability to balance efficiency with high performance. 
Further research explores automating network design through architectural search \cite{tan2019efficientnet, howard2019searching, tan2019mixconv} optimizing on the trade-off between accuracy and efficiency. 
In addition, network pruning techniques \cite{han2015deep} have been used to create efficient networks by removing unnecessary parts of larger networks, resulting in faster processing times and lower power consumption.

\subsection{Vision Transformer}
Transformers \cite{vaswani2017attention} have recently been applied to computer vision tasks, with the Vision Transformer (ViT) \cite{dosovitskiy2020image, touvron2021training} showing promising results in image classification.
However, the applicability of ViT has limitations for some downstream tasks due to the lack of downsampling and the high computational cost of self-attention.
To address these limitations, Pyramid Vision Transformer (PVT) \cite{wang2021pyramid} and Swin Transformer \cite{liu2021swin} were introduced, adapting the multi-scale encoder structures of convolutional neural networks to transformer-based hierarchical architectures. 
In addition to these two networks, there are several approaches \cite{wang2022pvt, hatamizadeh2022global} that attempt to achieve higher accuracy with better performance.
PLG-ViT \cite{ebert2023plg} stands out in their approach, combining local and global self-attention mechanisms in a parallel layer design. This parallel design ensures the computational efficiency of local-window attention while maintaining global information and achieving a high degree of accuracy. The global path is adaptive, applying self-attention to a fixed number of tokens, regardless of input resolution. While not optimized for devices with limited processing power, this unique property allows the network to achieve an optimal trade-off between complexity and performance, even with large resolution input, when compared to other SoTA vision transformer architectures.

\subsection{Light-Weight ViTs}
All of the above methods are not primarily designed for use on limited hardware, although there are some lightweight variations of some of these networks \cite{touvron2021training, wang2022pvt, hatamizadeh2022global}. 
These offshoots mainly try to save parameter by reducing the number of layers and channels.
However, there are first approaches of efficient on-device ViT architectures to port the performance of ViT to small hardware.
MobileViTs \cite{chen2022mobile, mehta2022separable, wadekar2022mobilevitv3} are the first ViT variant designed for mobile devices, but they are not as efficient as current SoTA CNNs like MobileNets and EfficientNets.
MobileFormer \cite{chen2022mobile} is able to deliver comparable or superior trade-offs in terms of accuracy and efficiency compared to both CNNs and existing ViTs, but requires a large number of parameters to achieve competitive performance.
By using of hierarchical patch-based embedding, attention-guided depthwise separable convolutions, and spatial attention pruning, EdgeViT \cite{pan2022edgevits} achieves SoTA performance.
However, EdgeViT faces a common issue among ViTs where its complexity directly correlates to input resolution, resulting in significant performance degradation when processing high-resolution images.

To close the gap in accuracy and speed between ViTs and their lighter derivatives, we propose our Light-Weight (LW) PLG-ViT.
The original PLG-ViT defines itself by a high accuracy with low computational overhead.
However, for limited resources like CPUs or mobile GPUs even PLG-ViT Tiny is still to parameter heavy and computational intensive.
Our redesign represents the natural evolution, making LW PLG-ViT a promising solution for real-world edge-devices.

\section{Method}
In this section, we describe our approach in redesigning the original PLG-ViT \cite{ebert2023plg} architecture, which aims to reduce the number of parameters and floating-point operations (FLOPs) without sacrificing accuracy.
We begin by describing the original design, identify computationally expensive operations, and then describe our novel methods and proposed changes.

\subsection{Original PLG-ViT}
The hierarchical encoder structure in the original PLG-ViT architecture is composed of five stages, with the first stage consisting of a convolutional layer that acts as an overlapping patch-embedding. 
This tokenization layer reduces the image resolution by a factor of $4$ and projects the input into a $C$-dimensional embedding space. The subsequent four transformer stages consist of the PLG self-attention (PLG-SA) and a convolutional feed forward network, each preceded by a layer normalization \cite{ba2016layer}. Each transformer stage is separated by convolutional downsampling, which halves the resolution of the features while doubling the depth.

The PLG-SA splits the input along the feature depth to generate both local and global features. 
By reducing the feature maps, complexity and parameters are saved, thus decreasing the number of calculations. 
Local features undergo window self-attention while global features are processed through an adaptive patch-sampling operation that combines max- and average pooling. 
This operation creates a single global window with a fixed token size that is used for computing self-attention.
The property of adaptive patch-sampling makes PLG-ViT computationally efficient with increasing input resolution.
After applying self-attention to the global features, a bilinear interpolation is used to restore the original resolution. 
Finally, the local and global features are concatenated, and an FFN follows the self-attention layer.

\subsection{Model Redesign}
The architecture of PLG-ViT is already characterized by resolution independent efficiency, which is also our decisive argument for a redesign targeting weaker hardware. 
However, some structures of PLG-ViT were not designed with hardware limited applications in mind. The first network part to be redesigned is convolutional patch-embedding, which is applied before each transformer stage. 
The original patch-embedding is a modified version of the Fused-MBConv layer \cite{tan2021efficientnetv2, hatamizadeh2022global} and can be described by
\begin{equation}\label{eq:pe_old}
\begin{split}
    z^{*} &= \text{PWConv}(\text{SE}_{\frac{1}{4}}(\text{GeLU}(\text{DWConv}_{3\times3}(z_{in})))), \\
    z_{out} &= \text{LN}(\text{SConv}_{3\times3}(z^{*} + z_{in})), \\
    \end{split}
\end{equation}
where LN, GeLU and SE refers to Layer Normalization, Gaussian Error Linear Units and a Squeeze and Excitation block \cite{hu2018squeeze}. 
$\text{SConv}_{3\times3}$ indicates $3 \times 3$ convolutional layer with a stride of $2$.
In our new and lighter redesign of patch-embedding the first part of the layer is kept with slight exceptions. 
Only the GeLU operation is replaced by the more performant SiLU operation \cite{elfwing2018sigmoid} and the reduction of the channels of the SE layer is increased from $\frac{1}{4}$ to $\frac{1}{8}$.
In the second part of Equation \ref{eq:pe_old} we replaced the strided convolution by a $1 \times 1$ layer to adjust the spatial (stride of $2$) and channel dimensions and a depthwise $3 \times 3$ layer to generate high-level features.
All these changes allow to save almost $80\%$ of the original parameters of the block.
Our new patch-embedding can be described as follows:
\begin{equation}\label{eq:pe_new}
\begin{split}
    z^{**} &= \text{PWConv}(\text{SE}_{\frac{1}{8}}(\text{SiLU}(\text{DWConv}_{3\times3}(z_{in})))), \\
    z_{out} &= \text{LN}(\text{DWConv}_{3\times3}(\text{PWConv}(z^{**} + z_{in}))), \\
    \end{split}
\end{equation}
%


\begin{figure}[t]
\centering
\includegraphics[width=0.8\linewidth]{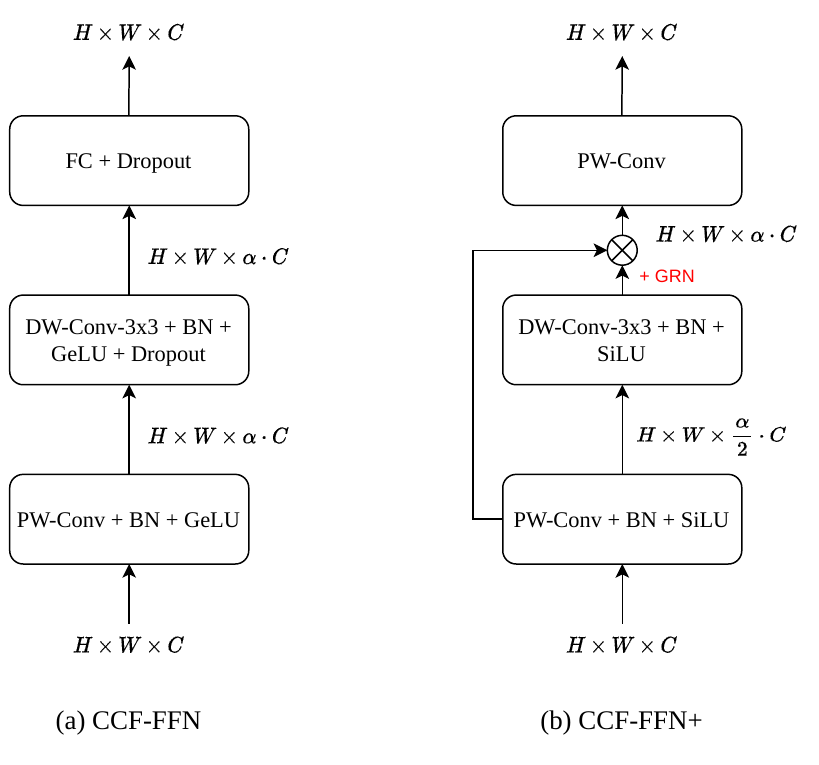}
\caption{Comparison of CCF-FFN \cite{ebert2023plg} and CCF-FFN+.}%
\label{fig:ffn}%
\end{figure}

Our second major change is in the convolutional feed forward network (FFN), which follows the PLG-SA layer as in Figure \ref{fig:architektur}. 
In Figure \ref{fig:ffn} (a) the CCF-FFN of the original PLG-ViT is visualized. 
At the beginning of the original FFN, the features are increased in depth by a factor $\alpha$, which is chosen as $4$ by default. 
This is followed by a depthwise $3 \times 3$ layer and a fully connected layer (FC), which restores the original depth. 
In the course of our redesign we add some new features to the CCF-FFN and generate our new CCF-FFN+ (see Figure \ref{fig:ffn} (b)), which on the one hand slightly reduces the number of parameters and on the other hand also increases the performance (see Section \ref{sec:ab}).
The general structure of the original FFN is preserved in our CCF-FFN+ and the features from the initial pointwise convolution are fed to the subsequent depthwise convolution.
However, these features are also concatenated with the features resulting from the depthwise $3 \times 3$ convolution. 
This fusion allows richer features while reducing the value of $\alpha$ from $4$ to $3$, which leads to a simultaneous reduction of the parameters.  
Furthermore, we utilize GRN (Global Response Normalization) of ConvNeXt-v2 \cite{woo2023convnext} right after the depthwise convolution to improve the feature representation capability of the model and reduce the computational cost of training.

Our last redesign targets patch-sampling in the global branch of the PLG-SA layer. 
The original sampling obtains a global window through added adaptive max- and average-pooling providing a fixed amount of tokens for the following global self-attention.
For larger network variants, this addition of pooled features was associated with an increase in accuracy. 
We observe that this does not hold true for our optimized variants, as such we find it beneficial to omit average pooling to reduce memory consumption and FLOPs.

\subsection{Architecture Variants} \label{sec:variants}

\begin{table}[t]
\caption{Model configurations of our LW PLG-ViT.  
}
\label{tab:config}
\centering
\begin{tabular}{cccccccc}
\hline
\textbf{}          & \textbf{Layer}          \hspace{-2.5mm} & \multicolumn{3}{c}{\textbf{LW PLG-ViT-A}}                                  \hspace{-2.5mm} & \multicolumn{3}{c}{\textbf{LW PLG-ViT-R}}                                 \\ \hline
0                  & CNN-Stem                \hspace{-2.5mm} & \multicolumn{3}{c}{$s=4, C1=64 $}                                          \hspace{-2.5mm} & \multicolumn{3}{c}{$s=4, C1=48 $}                                         \\ \hline
\multirow{2}{*}{1} & \multirow{2}{*}{LW PLG} \hspace{-2.5mm} & \multirow{2}{*}{$[$} & \hspace{-2.5mm}$lsa = \{7, 1\}$  & \hspace{-2.5mm}\multirow{2}{*}{$] \times 3 $} \hspace{-2.5mm} & \multirow{2}{*}{$[$} & \hspace{-2.5mm}$lsa = \{7, 1\}$  & \hspace{-2.5mm}\multirow{2}{*}{$] \times 1$} \\
                   &                         \hspace{-2.5mm} &                      & \hspace{-2.5mm}$gsa = \{7, 1\}$  &                                \hspace{-2.5mm} &                      & \hspace{-2.5mm}-                  &                               \\ \hline
\multirow{3}{*}{2} & CNN-Down                \hspace{-2.5mm} & \multicolumn{3}{c}{$s=2, C2=96$}                                           \hspace{-2.5mm} & \multicolumn{3}{c}{$s=2, C2=96$}                                          \\ \cline{2-8} 
                   & \multirow{2}{*}{LW PLG} \hspace{-2.5mm} & \multirow{2}{*}{$[$} & \hspace{-2.5mm}$lsa = \{7, 1\}$  & \hspace{-2.5mm}\multirow{2}{*}{$] \times 4$}  \hspace{-2.5mm} & \multirow{2}{*}{$[$} & \hspace{-2.5mm}$lsa = \{7, 1\}$  & \hspace{-2.5mm}\multirow{2}{*}{$] \times 1$} \\
                   &                         \hspace{-2.5mm} &                      & \hspace{-2.5mm}$gsa = \{14, 2\}$ &                                \hspace{-2.5mm} &                      & \hspace{-2.5mm}$gsa = \{14, 1\}$ &                               \\ \hline
\multirow{3}{*}{3} & CNN-Down                \hspace{-2.5mm} & \multicolumn{3}{c}{$s=2, C3=128$}                                          \hspace{-2.5mm} & \multicolumn{3}{c}{$s=2, C3=240$}                                         \\ \cline{2-8} 
                   & \multirow{2}{*}{LW PLG} \hspace{-2.5mm} & \multirow{2}{*}{$[$} & \hspace{-2.5mm}$lsa = \{7, 2\}$  & \hspace{-2.5mm}\multirow{2}{*}{$] \times 12$} \hspace{-2.5mm} & \multirow{2}{*}{$[$} & \hspace{-2.5mm}$lsa = \{7, 2\}$  & \hspace{-2.5mm}\multirow{2}{*}{$] \times 3$} \\
                   &                         \hspace{-2.5mm} &                      & \hspace{-2.5mm}$gsa = \{14, 2\}$ &                                \hspace{-2.5mm} &                      & \hspace{-2.5mm}$gsa = \{14, 3\}$ &                               \\ \hline
\multirow{4}{*}{4} & CNN-Down                \hspace{-2.5mm} & \multicolumn{3}{c}{$s=2, C4=192$}                                          \hspace{-2.5mm} & \multicolumn{3}{c}{$s=2, C4=384$}                                         \\ \cline{2-8} 
                   & \multirow{2}{*}{LW PLG} \hspace{-2.5mm} & \multirow{2}{*}{$[$} & \hspace{-2.5mm}$lsa = \{7, 3\}$  & \hspace{-2.5mm}\multirow{2}{*}{$] \times 4$}  \hspace{-2.5mm} & \multirow{2}{*}{$[$} & \hspace{-2.5mm}$lsa = \{7, 4\}$  & \hspace{-2.5mm}\multirow{2}{*}{$] \times 1$} \\
                   &                         \hspace{-2.5mm} &                      & \hspace{-2.5mm}$gsa = \{7, 3\}$  &                                \hspace{-2.5mm} &                      &\hspace{-2.5mm} $gsa = \{7, 4\}$  &                               \\ \cline{2-8} 
                   & CNN-Exp.                \hspace{-2.5mm} & \multicolumn{3}{c}{$s=1, C5=576$}                                        \hspace{-2.5mm} & \multicolumn{3}{c}{$s=1, C5=960$}                                       \\ \hline
\end{tabular}
\vspace{-2mm}
\end{table}

The original PLG-ViT proposes three variants, which differ in the number of parameters and FLOPs. 
Since our goal is to develop a smaller and also lighter variant, we take the smallest model, PLG-ViT Tiny, as a baseline.

First, the number of parameters will be reduced. 
This is primarily achieved by reducing the depth of the feature maps at each stage. 
PLG-ViT Tiny has a feature depth of $C_{1}=64$ in the first transformer stage which doubles with each following stage.
Due to the continuous doubling of the features the number of parameters increases rapidly. 
To counteract this, the features in our light-weight variant are not doubled, but can be set individually, similar to MobileNet \cite{howard2017mobilenets}.
At the the final stage, we add a further $3 \times 3$ layer for channel expansion for richer features.

In the original PLG-ViT the number of local and global heads is always the same. 
However, we observe that the global self-attention is more performant due to fixed global window. As a consequence we decided to decouple local and global self-attention and allow an unequal number of parallel heads.
Thus we can preferentially perform global self-attention by using more global heads.
This can also be seen in Figure \ref{fig:architektur}. 
The input of the LW PLG-SA has the shape $ z \in \mathbb{R}^ {H \times W \times C}$, where $H$ and $W$ indicate the spatial dimension of the features and $C$ refers to feature depth.
In contrast to the original implementation, we  split the input $z$ along the feature depth, and generate the local features $z_{l} \in \mathbb{R}^ {H \times W \times C*r}$ and the global features $z_{g} \in \mathbb{R}^ {H \times W \times C*(1-r)}$.
Here $r$ describes the ratio between local and global windows and can be set individually. 
In the original PLG-SA, $r$ is always set to $0.5$.

In this paper, we propose two network configurations: Light-Weight PLG-ViT-A, which is accuracy-optimized, and the runtime optimized Light-Weight PLG-ViT-R. 
Both networks have 5M parameters in total but differ in the number of FLOPs.
The configuration of each model can be found in Table \ref{tab:config}. 
In the Table, we list the stride $s$, the channel-depth $C$ and more information of the transformer-settings at each stage. 
Here $gsa$ and $lsa$ refers to global and local self-attention.
The first number of the curly bracket indicates the window-size and the second number refers to the number of heads.
The number behind the the square bracket indicates the number of block repetitions.

\section{Evaluation}

\begin{table}[t]
\centering
\caption{Comparison of Top-1 (in $\%$) classification accuracy on the ImageNet-1K \cite{deng2009imagenet} validation set. \textit{Param} refers to the number of parameters and \textit{FLOPs} are calculated at an input resolution of $224^2$. Inference is measured on the Intel i5-10400F processor (single core).}
\label{tab:imagenet}
\begin{tabular}{lcccc}
\hline
\textbf{Method}    & \textbf{Param} & \textbf{FLOPs} & \textbf{CPU (ms)} & \textbf{Top-1} \\ \hline
\multicolumn{5}{c}{\textit{Convolutional Neural Networks}}                                     \\ \hline
MobileNet-v1 \cite{howard2017mobilenets}       & 4.2 M          & 0.6 G          & 12.0 $\pm$ 0.3       & 70.6                \\
MobileNet-v2 \cite{sandler2018mobilenetv2}       & 3.4 M          & 0.3 G          & 15.5 $\pm$ 0.4       & 72.0                  \\
MobileNet-v2 (1.4) \cite{sandler2018mobilenetv2} & 6.9 M          & 0.6 G          & 25.0 $\pm$ 0.7       & 74.7                \\
ShuffleNet x1.5 \cite{zhang2018shufflenet}    & 3.4 M          & 0.3 G          & 10.8 $\pm$ 0.3       & 71.5                \\
ShuffleNet x2 \cite{zhang2018shufflenet}      & 5.4 M          & 0.5 G          & 16.7 $\pm$ 0.2       & 73.7                \\
EfficientNet-B0 \cite{tan2019efficientnet}   & 5.3 M          & 0.4 G          & 22.1 $\pm$ 0.5       & 77.1                \\ \hline
\multicolumn{5}{c}{\textit{Transformers}}                                                      \\ \hline
MobileViTv1-S \cite{mehta2022mobilevit}     & 5.6 M          & 2.0 G            & 43.7 $\pm$ 1.8       & 78.4                \\
MobileViTv2-1.0 \cite{mehta2022separable}    & 4.9 M          & 1.9 G          & 42.1 $\pm$ 3.6       & 78.1                \\
MobileViTv3-1.0 \cite{wadekar2022mobilevitv3}    & 5.1 M          & 1.9 G          & 39.3 $\pm$ 1.2       & 78.6                \\
MobileFormer96M \cite{chen2022mobile}  & 4.6 M          & 0.1 G          & 18.4 $\pm$ 0.5        & 72.8                \\
MobileFormer151M \cite{chen2022mobile} & 7.6 M          & 0.2 G          & 24.5 $\pm$ 0.7       & 75.2                \\
DeiT-Tiny \cite{touvron2021training}         & 5.7 M          & 1.3 G          & 29.4 $\pm$ 1.3       & 72.2                \\
T2T-ViT-10 \cite{yuan2021tokens}        & 5.9 M          & 1.5 G          & 29.6 $\pm$ 0.8       & 75.2                \\
PvTv2-B0 \cite{wang2022pvt}          & 3.4 M          & 0.6 G          & 20.3 $\pm$ 1.2       & 70.5                \\
PvTv2-B1 \cite{wang2022pvt}          & 14.0 M           & 2.1 G          & 57.1 $\pm$ 1.5       & 78.7                \\
EdgeNeXt-S  \cite{maaz2023edgenext}          & 5.6 M           & 1.0 G          & 33.0 $\pm$ 0.8       & 78.8                \\
EdgeViT-XXS \cite{pan2022edgevits}        & 4.1 M          & 0.6 G          & 27.3 $\pm$ 0.4       & 74.4                \\
EdgeViT-XS \cite{pan2022edgevits}         & 6.7 M          & 1.1 G          & 34.1 $\pm$ 0.6       & 77.5                \\ 
\rowcolor[HTML]{EFEFEF} 
LW PLG-R (ours)          & 5.0 M            & 0.7 G          & 32.4 $\pm$ 1.1       & 76.7                \\
\rowcolor[HTML]{EFEFEF} 
LW PLG-A (ours)           & 5.0 M            & 1.6 G          & 66.8 $\pm$ 1.4       & 79.5                  \\ \hline
\end{tabular}
\end{table}

\begin{table*}[t!]
\caption{Object detection results on five vision benchmarks \cite{lin2014microsoft, yu2020bdd100k, DiasDaCruz2020SVIRO, leibner2023gerald, majchrowska2021agar} from diverse domains. Comparison of our LW PLG-ViT-R and LW PLG-ViT-A wit SoTA CNNs \cite{he2016deep, sandler2018mobilenetv2, tan2019efficientnet} and Transformers \cite{ pan2022edgevits, wang2022pvt} using Mask- \cite{he2017mask} / Faster-RCNN \cite{Ren_2017} and RetinaNet \cite{lin2017focal}. All models are trained with $1 \times$ scheduler.}
\centering
\scriptsize
\label{tab:od}
\begin{tabular}{lccccccccccc}
\hline
\multicolumn{1}{l|}{}                                        & \multicolumn{4}{c|}{\textbf{RetinaNet\cite{lin2017focal}} }                                                                                              & \multicolumn{7}{c}{\textbf{Mask-RCNN \cite{he2017mask} \& Faster-RCNN \cite{Ren_2017}}}                                                                                                                                                                                                   \\
\multicolumn{1}{l|}{\multirow{-2}{*}{\textbf{Method}}}       & \textbf{Param}                & \textbf{mAP\textsuperscript{box}} & \multicolumn{1}{c}{\textbf{AP50\textsuperscript{box}}} & \multicolumn{1}{c|}{\textbf{AP75\textsuperscript{box}}}                & \textbf{Param}                & \multicolumn{1}{c}{\textbf{mAP\textsuperscript{box}}} & \multicolumn{1}{c}{\textbf{AP50\textsuperscript{box}}} & \multicolumn{1}{c}{\textbf{AP75\textsuperscript{box}}} & \multicolumn{1}{c}{\textbf{mAP\textsuperscript{mask}}} & \multicolumn{1}{c}{\textbf{AP50\textsuperscript{mask}}} & \multicolumn{1}{c}{\textbf{AP75\textsuperscript{mask}}} \\ \hline
\multicolumn{12}{c}{\textit{MS COCO \cite{lin2014microsoft} @ $1280 \times 800$} \textit{and 80 classes}}\\ \hline
\multicolumn{1}{l|}{ResNet18 \cite{he2016deep}}                                & 21.3 M                         & 31.8        & 49.6                              & \multicolumn{1}{l|}{33.6}                         & 31.2 M                         & 33.9                            & 54.0                              & 36.5                              & 30.8                            & 50.8                              & 32.7                              \\
\multicolumn{1}{l|}{MobileNet-v2 \cite{sandler2018mobilenetv2}}                            & 11.9 M                         & 31.3        & 49.0                              & \multicolumn{1}{l|}{33.0}                         & 21.7 M                         & 29.4                            & 48.1                              & 30.9                              & 27.1                            & 45.6                              & 28.4                              \\
\multicolumn{1}{l|}{EfficientNet-B0 \cite{tan2019efficientnet}}                         & 20.0 M                         & 39.7        & 59.1                              & \multicolumn{1}{l|}{42.7}                         & 29.6 M                         & 37.1                            & 57.2                              & 40.2                              & 33.8                            & 54.4                              & 35.9                              \\
\multicolumn{1}{l|}{PvTv2-B0 \cite{wang2022pvt}}                                & 13.0 M                         & 37.2        & 57.2                              & \multicolumn{1}{l|}{39.5}                         & 23.5 M                         & 38.2                            & 60.5                              & 40.7                              & 36.2                            & 57.8                              & 38.6                              \\
\multicolumn{1}{l|}{PvTv2-B1 \cite{wang2022pvt}}                                & 23.8 M                         & 41.2        & 61.9                              & \multicolumn{1}{l|}{43.9}                         & 33.7 M                         & 41.8                            & 64.3                              & 45.9                              & 38.8                            & 61.2                              & 41.6                              \\
\multicolumn{1}{l|}{EdgeViT-XXS \cite{pan2022edgevits}}                             & 13.1 M                         & 38.7        & 59.0                              & \multicolumn{1}{l|}{41.0}                         & 23.8 M                         & 39.9                            & 62.0                              & 43.1                              & 36.9                            & 59.0                              & 39.4                              \\
\multicolumn{1}{l|}{EdgeViT-XS \cite{pan2022edgevits}}                              & 22.0 M                         & 40.6        & 61.3                              & \multicolumn{1}{l|}{43.3}                         & 32.8 M                         & 41.4                            & 63.7                              & 45.0                              & 38.3                            & 60.9                              & 41.3                              \\
\rowcolor[HTML]{EFEFEF} 
\multicolumn{1}{l|}{\cellcolor[HTML]{EFEFEF}LW PLG-R (ours)} & 13.6 M                         & 38.0        & 58.8                              & \multicolumn{1}{l|}{\cellcolor[HTML]{EFEFEF}40.0} & 23.8 M                         & 38.5                            & 60.8                              & 41.7                              & 35.7                            & 57.8                              & 37.8                              \\
\rowcolor[HTML]{EFEFEF} 
\multicolumn{1}{l|}{\cellcolor[HTML]{EFEFEF}LW PLG-A (ours)} & 13.7 M                         & 40.8        & 61.5                              & \multicolumn{1}{l|}{\cellcolor[HTML]{EFEFEF}43.1} & 24.3 M                         & 41.5                            & 63.8                              & 44.9                              & 38.0                            & 60.5                              & 40.0                              \\ \hline
\multicolumn{12}{c}{\textit{BDD100K \cite{yu2020bdd100k} @ $1280 \times 720$} \textit{and 10 classes}} \\ \hline
\multicolumn{1}{l|}{ResNet18 \cite{he2016deep}}                                & 21.3 M                         & 26.1        & 48.8                              & \multicolumn{1}{l|}{23.8}                         & 28.5 M                         & 23.8                            & 45.4                              & 21.5                              & -                               & -                                 & -                                 \\
\multicolumn{1}{l|}{MobileNet-v2 \cite{sandler2018mobilenetv2}}                            & 11.9 M                         & 26.6        & 48.9                              & \multicolumn{1}{l|}{24.6}                         & 16.7 M                         & 21.4                            & 36.1                              & 21.5                              & -                               & -                                 & -                                 \\
\multicolumn{1}{l|}{EfficientNet-B0 \cite{tan2019efficientnet}}                         & 20.0 M                         & 28.3        & 51.8                              & \multicolumn{1}{l|}{26.3}                         & 26.6 M                         & 23.5                            & 41.4                              & 23.0                              & -                               & -                                 & -                                 \\
\multicolumn{1}{l|}{PvTv2-B0 \cite{wang2022pvt}}                                & 13.0 M                         & 30.4        & 55.6                              & \multicolumn{1}{l|}{28.3}                         & 20.4 M                         & 29.9                            & 55.3                              & 27.7                              & -                               & -                                 & -                                 \\
\multicolumn{1}{l|}{PvTv2-B1 \cite{wang2022pvt}}                                & 23.8 M                         & 31.6        & 57.3                              & \multicolumn{1}{l|}{29.9}                         & 30.7 M                         & 32.0                             & 58.3                               & 30.0                               & -                               & -                                 & -                                 \\
\rowcolor[HTML]{EFEFEF} 
\multicolumn{1}{l|}{\cellcolor[HTML]{EFEFEF}LW PLG-R (ours)} & \cellcolor[HTML]{EFEFEF}13.6 M & 30.8        & 56.4                              & \multicolumn{1}{l|}{\cellcolor[HTML]{EFEFEF}28.8} & 20.8 M                         & 31.0                            & 56.8                              & 29.1                              & -                               & -                                 & -                                 \\
\rowcolor[HTML]{EFEFEF} 
\multicolumn{1}{l|}{\cellcolor[HTML]{EFEFEF}LW PLG-A (ours)} & \cellcolor[HTML]{EFEFEF}13.7 M & 32.2        & 58.2                              & \multicolumn{1}{l|}{\cellcolor[HTML]{EFEFEF}30.3} & 21.3 M                         & 31.7                            & 57.1                              & 29.8                              & -                               & -                                 & -                                 \\ \hline
\multicolumn{12}{c}{\textit{SVIRO \cite{DiasDaCruz2020SVIRO} @ $960 \times 640$}  \textit{and 5 classes}}                                                                                                                                                                                                                                                                                                                                                                                                \\ \hline
\multicolumn{1}{l|}{ResNet18 \cite{he2016deep}}                                & 21.3 M                         & 48.1        & 67.7                              & \multicolumn{1}{l|}{54.3}                         & 28.5 M                         & 48.4                            & 69.1                              & 57.8                              & -                               & -                                 & -                                 \\
\multicolumn{1}{l|}{MobileNet-v2 \cite{sandler2018mobilenetv2}}                            & 11.9 M                         & 52.8        & 74.9                              & \multicolumn{1}{l|}{58.4}                         & 16.7 M                         & 47.5                            & 69.4                              & 56.2                              & -                               & -                                 & -                                 \\
\multicolumn{1}{l|}{EfficientNet-B0 \cite{tan2019efficientnet}}                         & 20.0 M                         & 59.4        & 80.1                              & \multicolumn{1}{l|}{68.7}                         & 26.6 M                         & 54.2                            & 76.0                              & 64.3                              & -                               & -                                 & -                                 \\
\multicolumn{1}{l|}{PvTv2-B0 \cite{wang2022pvt}}                                & 13.0 M                         & 41.3        & 55.8                              & \multicolumn{1}{l|}{44.4}                         & 20.4 M                         & 35.8                            & 49.4                              & 40.6                              & -                               & -                                 & -                                 \\
\multicolumn{1}{l|}{PvTv2-B1 \cite{wang2022pvt}}                                & 23.8 M                         & 45.2        & 62.1                              & \multicolumn{1}{l|}{49.0}                         & 30.7 M                         & 37.8                            & 51.0                              & 43.0                              & -                               & -                                 & -                                 \\
\rowcolor[HTML]{EFEFEF} 
\multicolumn{1}{l|}{\cellcolor[HTML]{EFEFEF}LW PLG-R (ours)} & \cellcolor[HTML]{EFEFEF}13.6 M & 39.1        & 54.8                              & \multicolumn{1}{l|}{\cellcolor[HTML]{EFEFEF}42.4} & \cellcolor[HTML]{EFEFEF}20.8 M & 42.7                            & 58.4                              & 49.9                              & -                               & -                                 & -                                 \\
\rowcolor[HTML]{EFEFEF} 
\multicolumn{1}{l|}{\cellcolor[HTML]{EFEFEF}LW PLG-A (ours)} & \cellcolor[HTML]{EFEFEF}13.7 M & 41.3        & 54.5                              & \multicolumn{1}{l|}{\cellcolor[HTML]{EFEFEF}45.9} & \cellcolor[HTML]{EFEFEF}21.3 M & 43.0                            & 59.7                              & 49.8                              & -                               & -                                 & -                                 \\ \hline
\multicolumn{12}{c}{\textit{GERALD \cite{leibner2023gerald} @ $1280 \times 720$} \textit{and 9 classes}}  \\ \hline
\multicolumn{1}{l|}{ResNet18 \cite{he2016deep}}                                & 21.3 M                         & 39.3        & 66.3                              & \multicolumn{1}{l|}{42.7}                         & 28.5 M                         & 44.8                            & 71.5                              & 50.9                              & -                               & -                                 & -                                 \\
\multicolumn{1}{l|}{MobileNet-v2 \cite{sandler2018mobilenetv2}}                            & 11.9 M                         & 27.8       & 48.3                              & \multicolumn{1}{l|}{28.5}                         & 16.7 M                         & 35.6                            & 54.8                              & 41.2                              & -                               & -                                 & -                                 \\
\multicolumn{1}{l|}{EfficientNet-B0 \cite{tan2019efficientnet}}                         & 20.0 M                         & 36.3        & 66.5                              & \multicolumn{1}{l|}{36.4}                         & 26.6 M                         & 38.5                            & 60.7                              & 44.5                              & -                               & -                                 & -                                 \\
\multicolumn{1}{l|}{PvTv2-B0 \cite{wang2022pvt}}                                & 13.0 M                         & 48.3       & 81.9                              & \multicolumn{1}{l|}{51.6}                         & 20.4 M                         & 47.5                            & 73.6                              & 54.3                              & -                               & -                                 & -                                 \\
\multicolumn{1}{l|}{PvTv2-B1 \cite{wang2022pvt}}                                & 23.8 M                         & 48.2        & 82.0                              & \multicolumn{1}{l|}{50.6}                         & 30.7 M                         & 49.1                           & 75.6                              & 56.1                              & -                               & -                                 & -                                 \\
\rowcolor[HTML]{EFEFEF} 
\multicolumn{1}{l|}{\cellcolor[HTML]{EFEFEF}LW PLG-R (ours)} & \cellcolor[HTML]{EFEFEF}13.6 M & 45.7       & 77.0                              & \multicolumn{1}{l|}{\cellcolor[HTML]{EFEFEF}49.1} & \cellcolor[HTML]{EFEFEF}20.8 M & 48.2                            & 74.7                              & 54.6                              & -                               & -                                 & -                                 \\
\rowcolor[HTML]{EFEFEF} 
\multicolumn{1}{l|}{\cellcolor[HTML]{EFEFEF}LW PLG-A (ours)} & \cellcolor[HTML]{EFEFEF}13.7 M & 48.4        & 80.6                              & \multicolumn{1}{l|}{\cellcolor[HTML]{EFEFEF}52.5} & \cellcolor[HTML]{EFEFEF}21.3 M & 49.6                            & 75.3                              & 57.1                              & -                               & -                                 & -                                 \\ \hline
\multicolumn{12}{c}{\textit{AGAR \cite{majchrowska2021agar} @ $1536 \times 1536$} \textit{and 5 classes}}                                                                                                                                                                                                                                                                                                                                                                                               \\ \hline
\multicolumn{1}{l|}{ResNet18 \cite{he2016deep}}                                & 21.3 M                         & 45.6        & 73.5                              & \multicolumn{1}{l|}{50.9}                         & 28.5 M                         & 53.1                            & 76.9                              & 62.6                              & -                               & -                                 & -                                 \\
\multicolumn{1}{l|}{MobileNet-v2 \cite{sandler2018mobilenetv2}}                            & 11.9 M                         & 51.3        & 77.5                              & \multicolumn{1}{l|}{58.8}                         & 16.7 M                         & 36.5                            & 53.4                              & 42.7                              & -                               & -                                 & -                                 \\
\multicolumn{1}{l|}{EfficientNet-B0 \cite{tan2019efficientnet}}                         & 20.0 M                         & 54.8        & 82.1                              & \multicolumn{1}{l|}{63.5}                         & 26.6 M                         & 41.4                            & 60.5                              & 49.4                              & -                               & -                                 & -                                 \\
\multicolumn{1}{l|}{PvTv2-B0 \cite{wang2022pvt}}                                & 13.0 M                         & 55.1        & 80.0                              & \multicolumn{1}{l|}{64.3}                         & 20.4 M                         & 55.8                            & 78.2                              & 66.3                              & -                               & -                                 & -                                 \\
\multicolumn{1}{l|}{PvTv2-B1 \cite{wang2022pvt}}                                & 23.8 M                         & 54.9        & 80.0                              & \multicolumn{1}{l|}{64.2}                         & 30.7 M                         & 55.8                            & 77.4                              & 66.8                              & -                               & -                                 & -                                 \\
\rowcolor[HTML]{EFEFEF} 
\multicolumn{1}{l|}{\cellcolor[HTML]{EFEFEF}LW PLG-R (ours)} & 13.6 M                         & 55.7         & 81.2                               & \multicolumn{1}{l|}{\cellcolor[HTML]{EFEFEF}65.1}  & 20.8 M                         & \cellcolor[HTML]{EFEFEF}55.7    & \cellcolor[HTML]{EFEFEF}77.6      & \cellcolor[HTML]{EFEFEF}66.7      & -                               & -                                 & -                                 \\
\rowcolor[HTML]{EFEFEF} 
\multicolumn{1}{l|}{\cellcolor[HTML]{EFEFEF}LW PLG-A (ours)} & 13.7 M                         & 56.8         & 81.6                               & \multicolumn{1}{l|}{\cellcolor[HTML]{EFEFEF}67.0}  & 21.3 M                         & \cellcolor[HTML]{EFEFEF}56.1    & \cellcolor[HTML]{EFEFEF}77.8      & \cellcolor[HTML]{EFEFEF}67.1      & -                               & -                                 & -                                 \\ \hline
\end{tabular}
\end{table*}

In this section, we present a comprehensive evaluation of our proposed network on various benchmarks.
We perform experiments on the ImageNet-1K \cite{deng2009imagenet} benchmark for image classification, COCO \cite{lin2014microsoft} and AGAR \cite{majchrowska2021agar} for general object detection and instance segmentation.
In addition we use various benchmarks in the autonomous driving and transporting, namely NuScenes \cite{caesar2020nuscenes}, BDD100k \cite{yu2020bdd100k} and SVIRO \cite{DiasDaCruz2020SVIRO}.
NuScense and BDD100k datasets primarily focus on the environment outside the vehicle, whereas the SVIRO dataset focuses on interior monitoring, which is equally essential task \cite{ebert2022multitask,da2021autoencoder}.
Moreover, we use GERALD \cite{leibner2023gerald} as benchmark for recognition of railroad traffic signs.
Our method was compared with the SoTA in the following evaluation, followed by an examination of the network components through an ablation study.
Some qualitative examples of our LW PLG-ViT on individual tasks are shown in Figure \ref{fig:cover}.

\subsection{Quantitative Evaluation}
\subsubsection{Image Classification}

To tackle the task of image classification, we employ the widely-used ImageNet-1K dataset which comprises more than 1 million images, spanning across 1K classes. 
Our classification approach combines global average pooling of the output features from the last transformer stage with a linear classifier. 
In our comparison, we list only methods which are of similar network size and complexity to allow a direct comparison. 
We use the exact same training strategies as the original network \cite{ebert2023plg}.

We present the results of our experiment on ImageNet-1K validation in Table \ref{tab:imagenet}.
Our Light-Weight PLG-ViT model demonstrates significant improvements in Top-1 accuracy, while having a similar number of parameters and model complexity (FLOPs) as most CNNs and ViTs. 
Notably, our accuracy optimized network LW PLG-ViT-A outperforms EfficientNet \cite{tan2019efficientnet} by $+2.4\%$ Top-1 accuracy. 
However, our results show that CNNs tend to have lower inference times on CPUs compared to ViTs due to their design.

Our approach also compares favorably to transformer-based networks. 
LW PLG-ViT-A achieves approximately $+1\%$ Top-1 accuracy with comparable parameters and $20\%$ fewer FLOPs compared to all three versions of MobileViT \cite{mehta2022mobilevit, mehta2022separable, wadekar2022mobilevitv3}. 
Although MobileFormer \cite{chen2022mobile} networks have low complexity and runtime, they require a higher number of parameters to achieve similar accuracy values. 
Even PVTv2 \cite{wang2022pvt} performs worse than our LW PLG-ViT-A ($-0.7\%$ Top-1 accuracy) despite having over twice the amount of parameters. 
Only EdgeNeXt \cite{maaz2023edgenext} and EdgeViT \cite{pan2022edgevits} show comparable results,
However, only our accuracy optimized network achieves an accuracy of over $79.0\%$.

In summary, our Light-Weight PLG-ViT model achieves SoTA results in terms of accuracy while maintaining a reasonable model complexity and parameter count.
We demonstrate that transformer-based models can achieve comparable accuracy to CNNs, albeit with slightly higher computational requirements. 
Our accuracy optimized network, LW PLG-ViT-A, outperforms several other notable models, including EfficientNet, MobileViT, MobileFormer, and PVTv2.

\subsubsection{Object Detection and Instance Segmentation}

To train and evaluate our model as a backbone for object detection and instance segmentation methods, we utilize the COCO dataset \cite{lin2014microsoft}. 
For object detection in the automotive domain, we evaluate on BDD100k \cite{yu2020bdd100k} and SVIRO \cite{DiasDaCruz2020SVIRO}.
Furthermore we include the GERALD \cite{leibner2023gerald} dataset for railroad signal detection. 
Finally, we utilize AGAR \cite{majchrowska2021agar} as dataset to showcase the performance of our networks on high-resolution images beyond autonomous driving. 
We implement our ImageNet-1K pre-trained models as feature extractors for standard detectors such as Faster-RCNN \cite{Ren_2017} and RetinaNet \cite{lin2017focal} for object detection and Mask-RCNN \cite{he2017mask} for added instance segmentation. 
We follow the standard $1\times$ training scheduler with 12 epochs, similar to the original PLG-ViT. 
The implementation of all methods is based on MMDetection \cite{chen2019mmdetection} and the results are shown in Table \ref{tab:od}.

We observe that our transformer-based method significantly outperforms CNNs on benchmarks using real image data, such as COCO, BDD100K, GERALD and AGAR. Interestingly, this observation does not hold true when considering the synthetic data of SVIRO where CNNs can perform better. 
A possible explanation for this phenomenon is the homogeneity of the models and scenes used in SVIRO, which can present a challenge for transformer-based architectures.


In direct comparison with EdgeViT on COCO, we observe similar accuracy between EdgeViT-XS and our LW PLG-ViT-A, but with $38\%$ and $26\%$ less parameters for RetinaNet and Mask-RCNN, respectively.
When considering PVTv2, our network variants surpass the smaller b0 variant, while our accuracy-optimized network achieves similar accuracy to the much larger PVTv2-b1.
In addition to COCO, our networks also show excellent performance on the other benchmarks in  head-to-head comparison with PVTv2. 
Especially on the SVIRO dataset with Faster-RCNN as detector our networks outperform PVTv2-b1 by more than $+5$ mAP.

Based on our findings, we conclude that transformer-based networks outperform traditional CNNs in challenging tasks such as object detection and instance segmentation for real-world scenarios.
Moreover, our two networks exhibit exceptional performance in these tasks, even when compared to much more parameter-intensive networks.
Some visual examples of results can be seen in Figure \ref{fig:vis} (a),(b),(c).

\subsubsection{3D Monocular Object Detection}

\begin{table}[t]
\caption{Comparison of our LW PLG-ViT backbones in comparison to ResNet for the task of monocular 3D object detection on NuScenes \cite{caesar2020nuscenes} using the FCOS3D \cite{wang2021fcos3d} detector.}
\centering
\label{tab:3dod}
\begin{tabularx}{0.48\textwidth}{Xcccc}
\hline
\textbf{Method}     & \textbf{Param} & \textbf{FLOPs} & \textbf{mAP}              & \textbf{NDS}             \\ \hline
ResNet50 \cite{he2016deep}            & 34.2 M        & 312 G          & 26.0                     & 31.7                     \\
ResNet101 \cite{he2016deep}          & 53.1 M              & 422 G              & 27.2                     & 32.9                     \\
ResNet101 + DCN \cite{he2016deep}    & 55.0 M        & 342 G          & 29.5 & 37.2 \\
\rowcolor[HTML]{EFEFEF} 
LW PLG-ViT-R (ours) & 13.6 M        & 214 G          & 24.4                     & 33.1                     \\
\rowcolor[HTML]{EFEFEF} 
LW PLG-ViT-A (ours) & 14.1 M               & 232 G                & 28.1                         &  36.3                        \\ \hline
\end{tabularx}
\end{table}

We also evaluate our networks on 3D monocular object detection using the NuScenes dataset \cite{caesar2020nuscenes}. Monocular object detection is particularly challenging due to the lack of 3D information in image data.
Our ImageNet-1K pre-trained models are used as backbones for FCOS3D \cite{wang2021fcos3d}. 
Furthermore, we follow the standard $1\times$ training scheduler with 12 epochs without any fine-tuning steps.
The implementation is based on MMDetection3D \cite{chen2019mmdetection}. Our results are shown in Table \ref{tab:3dod}.

\begin{figure*}[t]
\centering
\subfloat{\includegraphics[width = 0.9\linewidth]{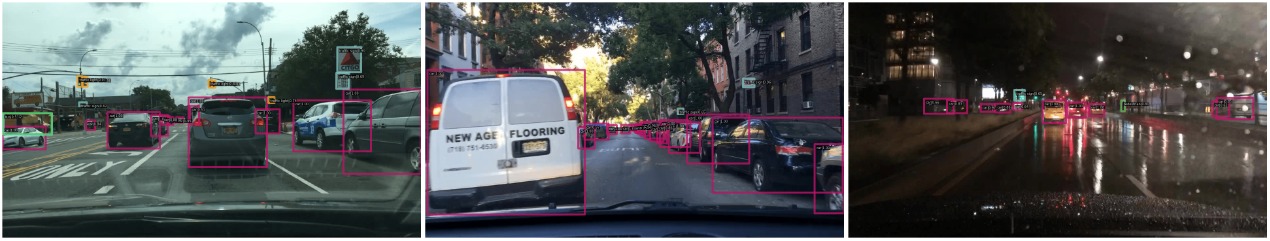}}\\ 
\subfloat{(a) 2D Object Detection Results on BDD100k \cite{yu2020bdd100k} with Faster-RCNN \cite{Ren_2017} and our LW PLG-ViT-A} \vspace{+1mm}\\

\subfloat{\includegraphics[width = 0.9\linewidth]{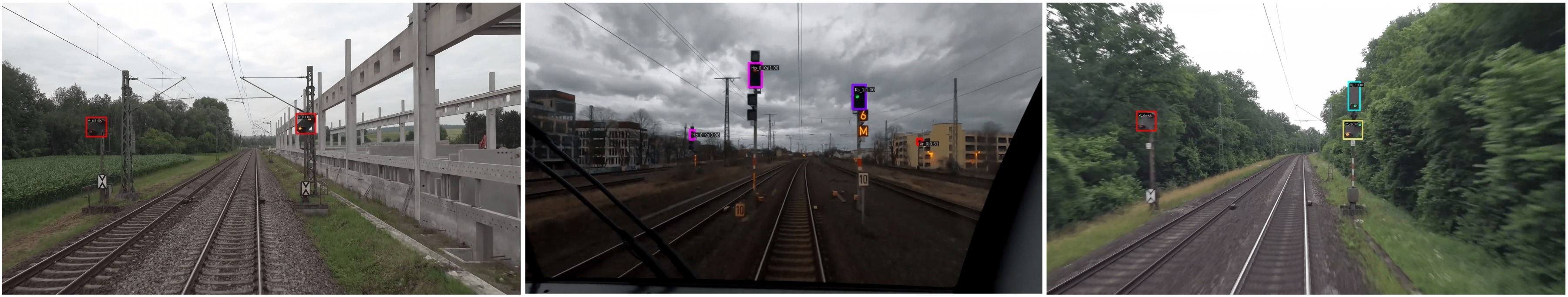}}\\ 
\subfloat{(b) 2D Object Detection Results on GERALD \cite{leibner2023gerald} with Faster-RCNN \cite{Ren_2017} and our LW PLG-ViT-A} \vspace{+1mm}\\

\subfloat{\includegraphics[width = 0.9\linewidth]{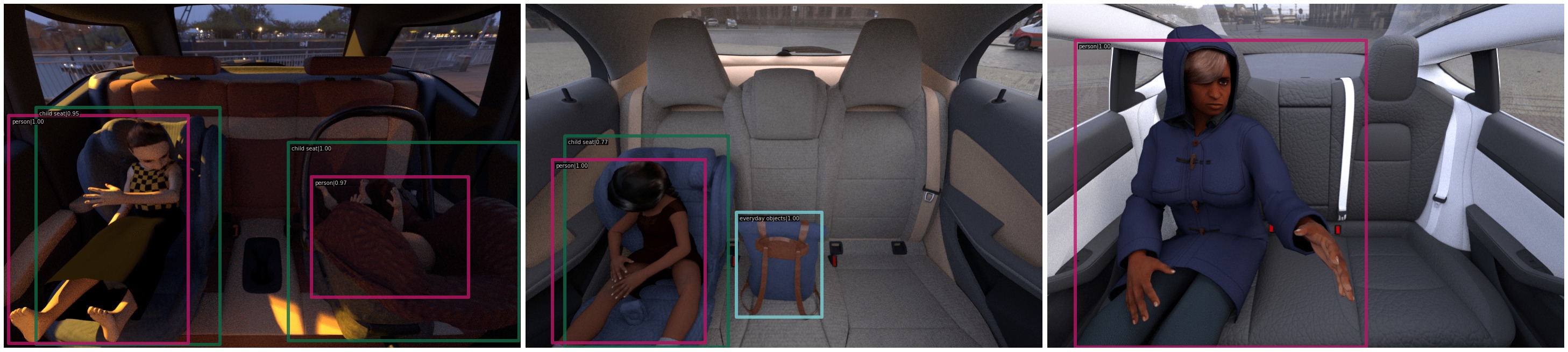}}\\ 
\subfloat{(c) 2D Object Detection Results on SVIRO \cite{DiasDaCruz2020SVIRO} with Faster-RCNN \cite{Ren_2017} and our LW PLG-ViT-A} \vspace{+1mm}\\

\subfloat{\includegraphics[width = 0.9\linewidth]{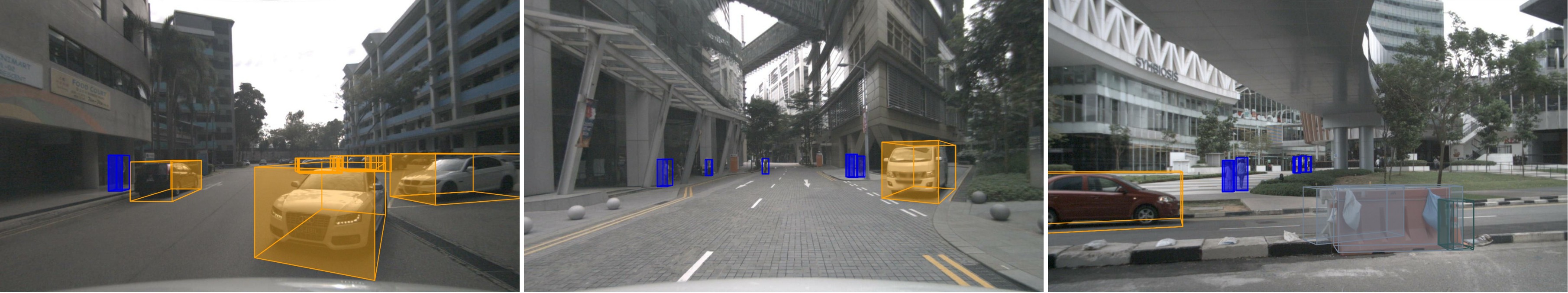}}\\ 
\subfloat{(d) 3D Monocular Object Detection Results on NuScenes \cite{caesar2020nuscenes} with FCOS3D \cite{wang2021fcos3d} and our LW PLG-ViT-A} \vspace{+1mm}\\

\caption{Visual Example detections of our Light-Weight PLG-ViT backbone on different datasets \cite{yu2020bdd100k, leibner2023gerald, DiasDaCruz2020SVIRO, caesar2020nuscenes}.}
\label{fig:vis}
\vspace{-4mm}
\end{figure*}

When compared with the much more parameter-intensive ResNet backbones our LW PLG-ViT shows on-par performance  despite significantly lower weights and complexity.
Our LW PLG-ViT-A even achieves a higher NuScense Detection Score (NDS) ($+4.6\%$ / $+3.4\%$) and mAP ($+2.1$ / $+0.9)$ than ResNet50 and ResNet101.
These results further verify the potential of transformer architecture. Some visual examples of our results can be seen in Figure \ref{fig:vis} (d).

\begin{figure}[b]
\begin{center}
\includegraphics[width=0.82\linewidth]{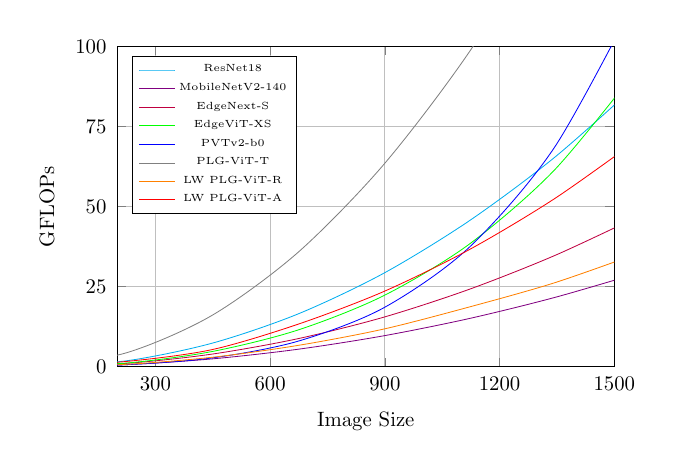}
\end{center}
  \caption{Complexity evaluation (GFLOPs) under different input sizes.}
\label{fig:complexity}
\end{figure}

\subsection{Ablation Study}\label{sec:ab}

\begin{table}[t]
\caption{Investigation of accuracy (in $\%$) influenced by new components of LW PLG-ViT on classification, using our LW PLG-VIT-A variant on the ImageNet-1K \cite{deng2009imagenet} dataset.}
\centering
\label{tab:ablation}
\begin{tabularx}{0.48\textwidth}{Xcccc}
\hline
\textbf{Method}   & \textbf{Param} & \textbf{FLOPs} & \textbf{Top-1} & \textbf{Top-5} \\ \hline
PLG-ViT-Tiny \cite{ebert2023plg}     & 26.6 M         & 4318 M         & 83.4           & 96.4           \\
+ Reparameterization        & 5.8 M          & 1619 M         & 79.0           & 94.6           \\
+ LW-Downsampling & 5.2 M          & 1452 M         & 78.6           & 94.7           \\
+ CCF-FFN+        & 5.0 M          & 1576 M         & 79.0           & 94.7           \\
+ Reduced patch-sampling         & 5.0 M          & 1576 M         & 79.0           & 94.7           \\
+ No Regularization         & 5.0 M          & 1576 M         & 79.5           & 95.3           \\ \hline
\end{tabularx}
\end{table}

In this section, we perform an ablation study on the crucial design elements and modules of our proposed model. 
The entire ablation study is presented using our accuracy optimized LW PLG-ViT-A.
For all experiments, we utilize the ImageNet-1K dataset and report the Top-1 and Top-5 accuracy on the validation set.
Table \ref{tab:ablation} reports the effectiveness of several components as part of the ablation study in comparison to the baseline of PLG-ViT Tiny.

In our first experiment, we perform a re-parametrization of the original PLG-ViT, decoupling local and global self-attention with the possibility of different numbers of heads for each branch. 
Accordingly to Section \ref{sec:variants}, we adjust the number of channels and layers and add a final expansion layer right in front of the classification head. 
These changes allow us to reduce the number of parameters from 26.6 M to only 5.8 M, which corresponds to a reduction of more than $80\%$.
In addition, this reduces the number of FLOPs over $60\%$.
These savings are accompanied by a deterioration in the top 1 accuracy of around $6\%$ from $83.4\%$ to $79.0\%$.
However, we deem this trade-off acceptable due to the immense savings in terms of parameters and complexity.
In our second experiment, we introduce our light-weight downsampling technique, which not only saves 0.6 M parameters but also reduces $10\%$ of the model's complexity in terms of FLOPS. 
However, this leads to a decrease in Top-1 accuracy by $-0.4\%$ from $79.0\%$ to $78.6\%$. 
We compensate this loss in accuracy by introducing our CCF-FFN+ module, which further reduces the number of parameters by an additional 0.2 M, altough with slight increase in FLOPs.
In the original PLG-ViT model, patch-sampling is crucial for achieving an increase of almost $+0.3\%$ Top-1 accuracy. 
However, for our light-weight redesign, we observe that simple max-pooling produces an equal Top-1 accuracy.
In our final experiment, we show that removing all regularization-methods (e.g. Dropout) increases the performance by almost $+0.5\%$ Top-1 accuracy.
Regularization is intended to prevent the overfitting of complex models to the training data.
However, since our network has so few parameters, regularization turns out to be counterproductive.

\subsection{Complexity Overhead Analysis}

Figure \ref{fig:complexity} illustrates the relationship between input size and the resulting growth in model complexity (in GFLOPS) for several models, including LW PLG-ViT, PVTv2-b0, EdgeNext-S, EdgeViT-XS, MobileNetV2, ResNet18 and the original PLG-ViT \cite{wang2022pvt, maaz2023edgenext, pan2022edgevits, sandler2018mobilenetv2, he2016deep, ebert2023plg}. The Figure demonstrates that as the input size increases, the LW PLG-ViT variants exhibit a significantly lower growth rate of GFLOPs compared to comparable ViT architectures. Notably, the standard self-attention of PVTv2 displays a quadratic increase in complexity with respect to the resolution. 
Furthermore, the complexity of both redesigns is also significantly lower than that of the original PLG-ViT. 
Overall, our networks demonstrate superior complexity characteristics, particularly for large image sizes.

\section{Conclusion}
In this paper, we presented our Light-Weight Parallel Local-Global Vision Transformer (LW PLG-ViT) as a  general-purpose backbone for image classification and dense downstream tasks.
Numerous redesigns enable us to optimize the already efficient PLG-ViT \cite{ebert2023plg} network to achieve high performance with comparatively weak hardware. 
This is particularly helpful for mobile applications such as autonomous driving.
We demonstrate the potential of our approach in various experiments and comparisons, outperforming different types of neural networks in multiple use cases.





\section*{ACKNOWLEDGMENT}

This research was partly funded by the Albert and Anneliese Konanz Foundation and the Federal Ministry of Education and Research Germany in the project M\textsuperscript{2}Aind-DeepLearning (13FH8I08IA).


\bibliographystyle{IEEEtran}
\bibliography{IEEEabrv,literatur}

\begin{thebibliography}{10}
\providecommand{\url}[1]{#1}
\csname url@samestyle\endcsname
\providecommand{\newblock}{\relax}
\providecommand{\bibinfo}[2]{#2}
\providecommand{\BIBentrySTDinterwordspacing}{\spaceskip=0pt\relax}
\providecommand{\BIBentryALTinterwordstretchfactor}{4}
\providecommand{\BIBentryALTinterwordspacing}{\spaceskip=\fontdimen2\font plus
\BIBentryALTinterwordstretchfactor\fontdimen3\font minus
  \fontdimen4\font\relax}
\providecommand{\BIBforeignlanguage}[2]{{%
\expandafter\ifx\csname l@#1\endcsname\relax
\typeout{** WARNING: IEEEtran.bst: No hyphenation pattern has been}%
\typeout{** loaded for the language `#1'. Using the pattern for}%
\typeout{** the default language instead.}%
\else
\language=\csname l@#1\endcsname
\fi
#2}}
\providecommand{\BIBdecl}{\relax}
\BIBdecl

\bibitem{dosovitskiy2020image}
A.~Dosovitskiy, L.~Beyer, A.~Kolesnikov, D.~Weissenborn, X.~Zhai,
  T.~Unterthiner, M.~Dehghani, M.~Minderer, G.~Heigold, S.~Gelly \emph{et~al.},
  ``An image is worth 16x16 words: Transformers for image recognition at
  scale,'' in \emph{International Conference on Learning Representations
  (ICLR)}, 2020.

\bibitem{liu2021swin}
Z.~Liu, Y.~Lin, Y.~Cao, H.~Hu, Y.~Wei, Z.~Zhang, S.~Lin, and B.~Guo, ``Swin
  transformer: Hierarchical vision transformer using shifted windows,'' in
  \emph{International Conference on Computer Vision (ICCV)}, 2021.

\bibitem{wang2021pyramid}
W.~Wang, E.~Xie, X.~Li, D.-P. Fan, K.~Song, D.~Liang, T.~Lu, P.~Luo, and
  L.~Shao, ``Pyramid vision transformer: A versatile backbone for dense
  prediction without convolutions,'' in \emph{International Conference on
  Computer Vision (ICCV)}, 2021.

\bibitem{he2016deep}
K.~He, X.~Zhang, S.~Ren, and J.~Sun, ``Deep residual learning for image
  recognition,'' in \emph{Conference on Computer Vision and Pattern Recognition
  (CVPR)}, 2016.

\bibitem{vaswani2017attention}
A.~Vaswani, N.~Shazeer, N.~Parmar, J.~Uszkoreit, L.~Jones, A.~N. Gomez,
  {\L}.~Kaiser, and I.~Polosukhin, ``Attention is all you need,'' \emph{Neural
  Information Processing Systems (NeurIPS)}, 2017.

\bibitem{Ren_2017}
S.~Ren, K.~He, R.~Girshick, and J.~Sun, ``Faster r-cnn: Towards real-time
  object detection with region proposal networks,'' \emph{Transactions on
  Pattern Analysis and Machine Intelligence (TPAMI)}, 2017.

\bibitem{yu2020bdd100k}
F.~Yu, H.~Chen, X.~Wang, W.~Xian, Y.~Chen, F.~Liu, V.~Madhavan, and T.~Darrell,
  ``Bdd100k: A diverse driving dataset for heterogeneous multitask learning,''
  in \emph{Conference on Computer Vision and Pattern Recognition (CVPR)}, 2020.

\bibitem{leibner2023gerald}
P.~Leibner, F.~Hampel, and C.~Schindler, ``Gerald: A novel dataset for the
  detection of german mainline railway signals,'' \emph{Proceedings of the
  Institution of Mechanical Engineers, Part F: Journal of Rail and Rapid
  Transit}, 2023.

\bibitem{mehta2022mobilevit}
S.~Mehta and M.~Rastegari, ``Mobilevit: Light-weight, general-purpose, and
  mobile-friendly vision transformer,'' in \emph{International Conference on
  Learning Representations (ICLR)}, 2022.

\bibitem{pan2022edgevits}
J.~Pan, A.~Bulat, F.~Tan, X.~Zhu, L.~Dudziak, H.~Li, G.~Tzimiropoulos, and
  B.~Martinez, ``Edgevits: Competing light-weight cnns on mobile devices with
  vision transformers,'' in \emph{European Conference on Computer Vision
  (ECCV)}, 2022.

\bibitem{mehta2022separable}
S.~Mehta and M.~Rastegari, ``Separable self-attention for mobile vision
  transformers,'' \emph{arXiv preprint arXiv:2206.02680}, 2022.

\bibitem{wadekar2022mobilevitv3}
S.~N. Wadekar and A.~Chaurasia, ``Mobilevitv3: Mobile-friendly vision
  transformer with simple and effective fusion of local, global and input
  features,'' \emph{arXiv preprint arXiv:2209.15159}, 2022.

\bibitem{ebert2023plg}
N.~Ebert, D.~Stricker, and O.~Wasenm{\"u}ller, ``Plg-vit: Vision transformer
  with parallel local and global self-attention,'' \emph{Sensors}, 2023.

\bibitem{deng2009imagenet}
J.~Deng, W.~Dong, R.~Socher, L.-J. Li, K.~Li, and L.~Fei-Fei, ``Imagenet: A
  large-scale hierarchical image database,'' in \emph{Conference on Computer
  Vision and Pattern Recognition (CVPR)}, 2009.

\bibitem{lin2014microsoft}
T.-Y. Lin, M.~Maire, S.~Belongie, J.~Hays, P.~Perona, D.~Ramanan,
  P.~Doll{\'a}r, and C.~L. Zitnick, ``Microsoft coco: Common objects in
  context,'' in \emph{European Conference on Computer Vision (ECCV)}, 2014.

\bibitem{caesar2020nuscenes}
H.~Caesar, V.~Bankiti, A.~H. Lang, S.~Vora, V.~E. Liong, Q.~Xu, A.~Krishnan,
  Y.~Pan, G.~Baldan, and O.~Beijbom, ``nuscenes: A multimodal dataset for
  autonomous driving,'' in \emph{Conference on Computer Vision and Pattern
  Recognition (CVPR)}, 2020.

\bibitem{DiasDaCruz2020SVIRO}
S.~{Dias Da Cruz}, O.~Wasenm\"uller, H.-P. Beise, T.~Stifter, and D.~Stricker,
  ``Sviro: Synthetic vehicle interior rear seat occupancy dataset and
  benchmark,'' in \emph{Winter Conference on Applications of Computer Vision
  (WACV)}, 2020.

\bibitem{howard2017mobilenets}
A.~G. Howard, M.~Zhu, B.~Chen, D.~Kalenichenko, W.~Wang, T.~Weyand,
  M.~Andreetto, and H.~Adam, ``Mobilenets: Efficient convolutional neural
  networks for mobile vision applications,'' \emph{arXiv preprint
  arXiv:1704.04861}, 2017.

\bibitem{sandler2018mobilenetv2}
M.~Sandler, A.~Howard, M.~Zhu, A.~Zhmoginov, and L.-C. Chen, ``Mobilenetv2:
  Inverted residuals and linear bottlenecks,'' in \emph{Conference on Computer
  Vision and Pattern Recognition (CVPR)}, 2018.

\bibitem{zhang2018shufflenet}
X.~Zhang, X.~Zhou, M.~Lin, and J.~Sun, ``Shufflenet: An extremely efficient
  convolutional neural network for mobile devices,'' in \emph{Conference on
  Computer Vision and Pattern Recognition (CVPR)}, 2018.

\bibitem{siam2018comparative}
M.~Siam, M.~Gamal, M.~Abdel-Razek, S.~Yogamani, M.~Jagersand, and H.~Zhang, ``A
  comparative study of real-time semantic segmentation for autonomous
  driving,'' in \emph{Conference on Computer Vision and Pattern Recognition
  Workshops (CVPRW)}, 2018.

\bibitem{tan2019efficientnet}
M.~Tan and Q.~Le, ``Efficientnet: Rethinking model scaling for convolutional
  neural networks,'' in \emph{International Conference on Machine Learning
  (ICML)}.\hskip 1em plus 0.5em minus 0.4em\relax PMLR, 2019.

\bibitem{howard2019searching}
A.~Howard, M.~Sandler, G.~Chu, L.-C. Chen, B.~Chen, M.~Tan, W.~Wang, Y.~Zhu,
  R.~Pang, V.~Vasudevan \emph{et~al.}, ``Searching for mobilenetv3,'' in
  \emph{International Conference on Computer Vision (ICCV)}, 2019.

\bibitem{tan2019mixconv}
M.~Tan and Q.~V. Le, ``Mixconv: Mixed depthwise convolutional kernels,''
  \emph{arXiv preprint arXiv:1907.09595}, 2019.

\bibitem{han2015deep}
S.~Han, H.~Mao, and W.~J. Dally, ``Deep compression: Compressing deep neural
  networks with pruning, trained quantization and huffman coding,'' in
  \emph{International Conference on Learning Representations (ICLR)}, 2016.

\bibitem{touvron2021training}
H.~Touvron, M.~Cord, M.~Douze, F.~Massa, A.~Sablayrolles, and H.~J{\'e}gou,
  ``Training data-efficient image transformers \& distillation through
  attention,'' in \emph{International Conference on Machine Learning (ICML)},
  2021.

\bibitem{wang2022pvt}
W.~Wang, E.~Xie, X.~Li, D.-P. Fan, K.~Song, D.~Liang, T.~Lu, P.~Luo, and
  L.~Shao, ``Pvt v2: Improved baselines with pyramid vision transformer,''
  \emph{Computational Visual Media}, 2022.

\bibitem{hatamizadeh2022global}
A.~Hatamizadeh, H.~Yin, J.~Kautz, and P.~Molchanov, ``Global context vision
  transformers,'' \emph{arXiv preprint arXiv:2206.09959}, 2022.

\bibitem{chen2022mobile}
Y.~Chen, X.~Dai, D.~Chen, M.~Liu, X.~Dong, L.~Yuan, and Z.~Liu,
  ``Mobile-former: Bridging mobilenet and transformer,'' in \emph{Conference on
  Computer Vision and Pattern Recognition (CVPR)}, 2022.

\bibitem{ba2016layer}
J.~L. Ba, J.~R. Kiros, and G.~E. Hinton, ``Layer normalization,'' \emph{arXiv
  preprint arXiv:1607.06450}, 2016.

\bibitem{tan2021efficientnetv2}
M.~Tan and Q.~Le, ``Efficientnetv2: Smaller models and faster training,'' in
  \emph{International Conference on Machine Learning (ICML)}, 2021.

\bibitem{hu2018squeeze}
J.~Hu, L.~Shen, and G.~Sun, ``Squeeze-and-excitation networks,'' in
  \emph{Conference on Computer Vision and Pattern Recognition (CVPR)}, 2018.

\bibitem{elfwing2018sigmoid}
S.~Elfwing, E.~Uchibe, and K.~Doya, ``Sigmoid-weighted linear units for neural
  network function approximation in reinforcement learning,'' \emph{Neural
  Networks}, 2018.

\bibitem{woo2023convnext}
S.~Woo, S.~Debnath, R.~Hu, X.~Chen, Z.~Liu, I.~S. Kweon, and S.~Xie, ``Convnext
  v2: Co-designing and scaling convnets with masked autoencoders,'' \emph{arXiv
  preprint arXiv:2301.00808}, 2023.

\bibitem{yuan2021tokens}
L.~Yuan, Y.~Chen, T.~Wang, W.~Yu, Y.~Shi, Z.-H. Jiang, F.~E. Tay, J.~Feng, and
  S.~Yan, ``Tokens-to-token vit: Training vision transformers from scratch on
  imagenet,'' in \emph{Conference on Computer Vision and Pattern Recognition
  (CVPR)}, 2021.

\bibitem{maaz2023edgenext}
M.~Maaz, A.~Shaker, H.~Cholakkal, S.~Khan, S.~W. Zamir, R.~M. Anwer, and
  F.~Shahbaz~Khan, ``Edgenext: efficiently amalgamated cnn-transformer
  architecture for mobile vision applications,'' in \emph{European Conference
  on Computer Vision Workshops (ECCVW)}, 2022.

\bibitem{majchrowska2021agar}
S.~Majchrowska, J.~Paw{\l}owski, G.~Gu{\l}a, T.~Bonus, A.~Hanas, A.~Loch,
  A.~Pawlak, J.~Roszkowiak, T.~Golan, and Z.~Drulis-Kawa, ``Agar a microbial
  colony dataset for deep learning detection,'' \emph{arXiv preprint
  arXiv:2108.01234}, 2021.

\bibitem{he2017mask}
K.~He, G.~Gkioxari, P.~Doll{\'a}r, and R.~Girshick, ``Mask r-cnn,'' in
  \emph{International Conference on Computer Vision}, 2017.

\bibitem{lin2017focal}
T.-Y. Lin, P.~Goyal, R.~Girshick, K.~He, and P.~Doll{\'a}r, ``Focal loss for
  dense object detection,'' in \emph{International Conference on Computer
  Vision (ICCV)}, 2017.

\bibitem{ebert2022multitask}
N.~Ebert, P.~Mangat, and O.~Wasenm{\"u}ller, ``Multitask network for joint
  object detection, semantic segmentation and human pose estimation in vehicle
  occupancy monitoring,'' in \emph{Intelligent Vehicles Symposium (IV)}, 2022.

\bibitem{da2021autoencoder}
S.~D. Da~Cruz, B.~Taetz, O.~Wasenm{\"u}ller, T.~Stifter, and D.~Stricker,
  ``Autoencoder based inter-vehicle generalization for in-cabin occupant
  classification,'' in \emph{Intelligent Vehicles Symposium (IV)}, 2021.

\bibitem{chen2019mmdetection}
K.~Chen, J.~Wang, J.~Pang, Y.~Cao, Y.~Xiong, X.~Li, S.~Sun, W.~Feng, Z.~Liu,
  J.~Xu \emph{et~al.}, ``Mmdetection: Open mmlab detection toolbox and
  benchmark,'' \emph{arXiv preprint arXiv:1906.07155}, 2019.

\bibitem{wang2021fcos3d}
T.~Wang, X.~Zhu, J.~Pang, and D.~Lin, ``Fcos3d: Fully convolutional one-stage
  monocular 3d object detection,'' in \emph{International Conference on
  Computer Vision (ICCV)}, 2021.

\end{thebibliography}

\end{document}